\documentclass[a4paper,twocolumn,twoside,10pt]{ranlp}

\usepackage{times}
\usepackage{epsfig}

\newif\ifpdf
\ifx\pdfoutput\undefined \else
  \ifx\pdfoutput\relax
  \else
    \ifcase\pdfoutput
    \else
      \pdftrue
    \fi
  \fi
\fi

\usepackage{abbrevs}
\let\em\itswitch

\newabbrev\tdt{Topic Detection and Tracking (TDT)}[TDT]
\newabbrev\mds{Multi-document Summarization}[MDS]
\newabbrev\sds{Single Document Summarization (SDS)}[SDS]
\newabbrev\sdrs{Synchronic and Diachronic Relations}[SDRs]
\newabbrev\RST{Rhetorical Structure Theory}[RST]
\newabbrev\cst{Cross-do\-cu\-ment Structure Theory (CST)}[CST]
\newabbrev\NLG{Natural Language Generation (NLG)}[NLG]
\newabbrev\IE{Information Extraction (IE)}[IE]
\newabbrev\nerc{Named Entity Recognition and Classification (NERC)}[NERC]
\newabbrev\nes{Named Entities (NEs)}[NEs]
\newabbrev\muc{Message Understanding Conferences (MUC)}[MUC]
\newabbrev\ml{Machine Learning}[ML]

\begin{document}

\title{Some Reflections on the Task of Content Determination in the Context of Multi-Document Summarization of Evolving Events}
\author{Stergos D. Afantenos}
\date{\small{Laboratoire d'Informatique Fondamentale de Marseille}\\
\small{Centre National de la Recherche Scientific (LIF - CNRS - UMR 6166)}\\
\small{Universit\'e de la M\'editerran\'ee, Facult\'e des Sciences de Luminy}\\
\small{163, Avenue de Luminy - Case 901, 13288 Marseille C\'edex 9 - France}\\
\texttt{stergos.afantenos@lif.univ-mrs.fr}}

\maketitle

\begin{abstract}
Despite its importance, the task of summarizing evolving events has received small attention by researchers in the field of \mds. In a previous paper \cite{Afantenos&al.07:JIIS} we have presented a methodology for the automatic summarization of documents, emitted by multiple sources, which describe the evolution of an event. At the heart of this methodology lies the identification of similarities and differences between the various documents, in two axes: the synchronic and the diachronic. This is achieved by the introduction of the notion of \emph{Synchronic and Diachronic Relations}. Those relations connect the messages that are found in the documents, resulting thus in a graph which we call \emph{grid}. Although the creation of the grid completes the Document Planning phase of a typical NLG architecture, it can be the case that the number of messages contained in a grid is very large, exceeding thus the required compression rate. In this paper we provide some initial thoughts on a probabilistic model which can be applied at the Content Determination stage, and which tries to alleviate this problem.

\bigskip\noindent\textit{\textbf{Keywords~:} summarization of evolving events, multi-document summarization, natural language generation}
\end{abstract}

\ResetAbbrevs{All}

\section{Introduction}
It wouldn't be an exaggeration to claim that human beings live engulfed in an environment full of information. Information which, metaphorically speaking, vie with each other in order to gain our attention, to gain an almost exclusive control of the precious resources which are our brains. This is most evident in the medium of Internet in which so many people are spending nowadays a considerable amount of their time. Information in this medium is constantly flowing in front of our screens, making the assimilation of such a plethora no longer feasible. In such an environment, information which is presented in brief and concise manner---\emph{i.e.} summarized information---stand more chances of retaining our attention, in relation to information presented in long and fragmented pieces of text. We can claim then, with a certain degree of certainty, that the task of automatic text summarization can prove to be very useful.

To provide a concrete example, we can imagine the case of a person who would like to keep track of the information related to an event as the event is evolving through time. What will usually happen in such cases is that, firstly, there will be more than one sources which will provide an account of the event, and secondly, most of the sources will provide more than one descriptions, in the sense that they will most probably follow the evolution of the event and provide updates as the event evolves through time. This can easily result in hundreds or even thousands of related articles which will describe the evolution of the same event, rendering it thus almost impossible for the interested person to read through its evolution comparing along the way the points in which the sources agree, disagree or present the information from a different point of view. A simple visit to a news aggregator, such as for example Google News,\footnote{\url{http://news.google.com/}} can make this point very clear.

As we have hinted before, a solution to this problem might be the automatic creation of summaries. In this paper we will present a methodology which aims at exactly that, \emph{i.e.}  the automatic creation of text summaries from documents emitted by multiple sources which describe the evolution of a particular event. In Section~\ref{sec:PhD} we will briefly present this methodology, at the heart of which lies the notion of \emph{\sdrs}(\sdrs) whose aim is the identification of the similarities and differences that exist between the documents in the synchronic and diachronic axes. The end result of this methodology is a graph whose vertices are the \sdrs and whose nodes are some structures which we call \emph{messages}. The creation of this graph can be considered as completing---as we have previously argued \cite{Afantenos&al.07:JIIS}---the \emph{Document Planning} phase of a typical architecture of a \NLG system \cite{Reiter&Dale2000}. Nevertheless, this graph can prove to be very large and thus the resulting summary can easily exceed the desired compression rate. In Section~\ref{sec:model} we will present a brief sketch of a probabilistic model for the selection of the appropriate information---\emph{i.e.} messages---to be included in the final summary, so that the desired compression rate will not be violated. In other words, we will propose a model for the \emph{Content Determination} stage of the Document Planning phase. This model will be based on certain remarks concerning the way with which information overlap between multiple documents which we present in Section~\ref{sec:greyareas}. The conclusions of this paper are presented in Section~\ref{sec:concl}.

\section{A Methodology for Summarizing Evolving Events\protect\footnote{Due to space limitations this section contains a very brief introduction to a methodology for the creation of summaries from evolving events that we have earlier presented \protect\cite{Afantenos&al.07:JIIS}. The interested reader is encouraged to consult \protect\cite{Afantenos.06:PhD:English,Afantenos&al.04:SETN,Afantenos&al.05:RANLP,Afantenos&al.07:JIIS,Afantenos&al.05:NLUCS} for more information.}}\label{sec:PhD}
At the heart of \mdslong(MDS) lies the process of identifying the similarities and differences that exist between the input documents. Although this holds true for the general case of \mds, for the case of summarizing \emph{evolving events} the identification of the similarities and differences should be distinguished, as we have previously argued \cite{Afantenos.06:PhD:English,Afantenos&al.04:SETN,Afantenos&al.05:RANLP,Afantenos&al.07:JIIS,Afantenos&al.05:NLUCS} between two axes: the \emph{synchronic} and the \emph{diachronic} axes. In the synchronic axis we are mostly concerned with the degree of agreement or disagreement that the various sources exhibit, for the same time frame, whilst in the diachronic axis we are concerned with the actual evolution of an event, as this evolution is being described by one source.

The initial inspiration for the \sdrsshort was provided by the \emph{\RSTlong}(RST) of Mann \& Thompson \cite{Mann&Thompson87,Mann&Thompson88}. \RST---which was initially developed in the context of ``computational text generation''\footnote {Also referred to as \NLGlong.} \cite{Mann&Thompson87,Mann&Thompson88,Taboada&Mann:RST1}---is trying to connect several \emph{units of analysis} with relations that are semantic in nature and are supposed to capture the intentions of the author. As ``units of analysis'' today are used, almost ubiquitously, the clauses of the text. In our case, as units of analysis for the \sdrs we are using some structures which we call \emph{messages}, inspired from the research in the \NLG field. Each message is composed of two parts: its \emph{type} and a list of \emph{arguments} which take their values from an \emph{ontology} for the specific domain. In other words, a message can be defined as follows:
\begin{center}
  \texttt{message\_type ( arg$_1$, $\ldots$ , arg$_n$ )}\\
  where \texttt{arg}$_i$ $\in$ Domain Ontology
\end{center}
The message type represents the type of the action that is involved in an event, whilst the arguments represent the main entities that are involved in this action. Additionally, each message is accompanied by information on the source which emitted this message, as well as its publication and referring time.

Concerning the \sdrs, in order to formally define a relation the following four fields ought to be defined (see also \cite{Afantenos&al.07:JIIS}):
\begin{enumerate}
  \item The relation's type (\emph{i.e.} Synchronic or Diachronic).
  \item The relation's name.
  \item The set of pairs of message types that are involved in the relation.
  \item The constraints that the corresponding arguments of each of the pairs
        of message types should have. Those constraints are expressed using the
        notation of first order logic.
\end{enumerate}
The name of the relation carries \emph{semantic} information which, along with the messages that are connected with the relation, are later being exploited by the \NLG component (see \cite{Afantenos&al.07:JIIS})
in order to produce the final summary.

\begin{figure}[htb]
\begin{center}
  \ifpdf
    \includegraphics[width=\linewidth]{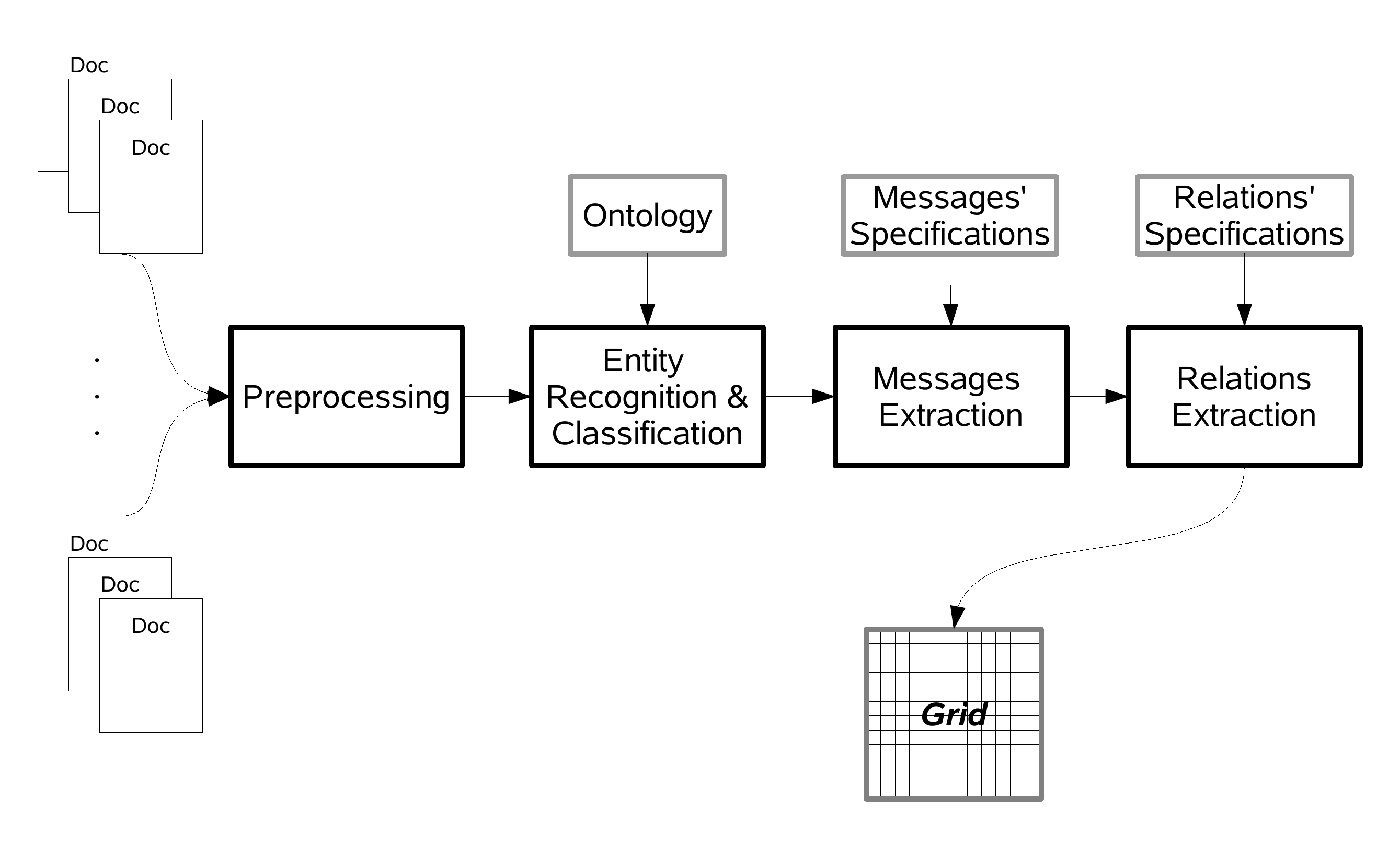}
  \else
    \includegraphics[width=\linewidth]{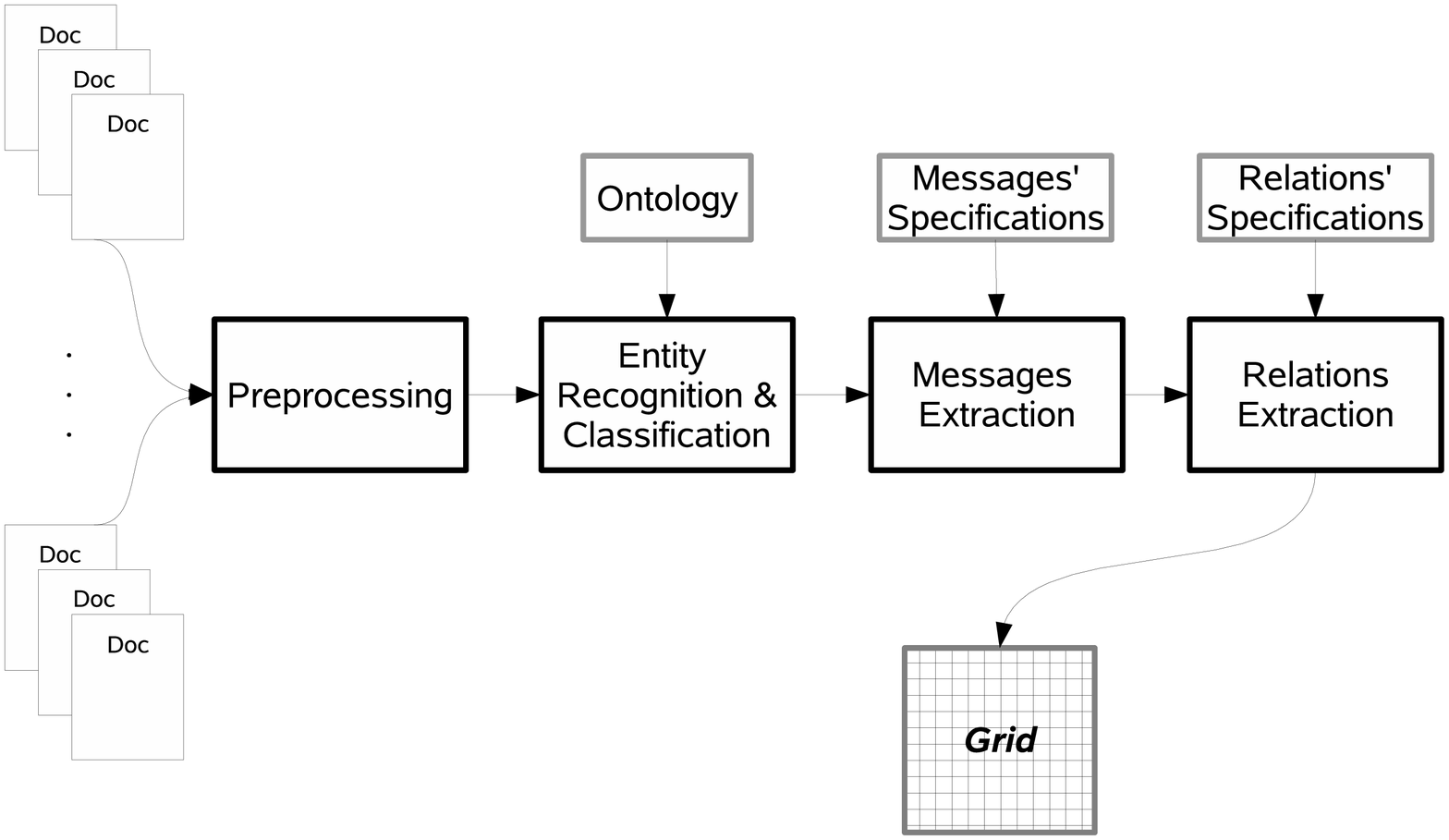}
  \fi
  \caption{The summarization system.}\label{fig:summarization_core}
\end{center}
\end{figure}

The methodology we propose consists of two main phases, the \emph{topic analysis phase} and the \emph{implementation phase}. The topic analysis phase is composed of four steps, which include the creation of the ontology for the topic and the providing of the specifications for the messages and the \sdrs. The final step of this phase, which in fact serves as a bridge step with the implementation phase, includes the annotation of the corpora belonging to the topic under examination that have to be collected as a preliminary step during this phase. The annotated corpora will serve a dual role: the first is the training of the various \ml algorithms used during the next phase and the second is for evaluation purposes. The implementation phase
involves the computational extraction of the messages and the \sdrs that connect them in order to create a directed acyclic graph (DAG) which we call \emph{grid}. The architecture of the summarization system is shown in Figure~\ref{fig:summarization_core}.

We applied our methodology in two different case studies. The first case study concerned the description of football matches, a topic which evolved linearly and exhibited synchronous emission of reports, while the second case study concerned the description of terroristic incidents with hostages, a topic which evolved non-linearly and exhibited asynchronous emission of reports.\footnote{On the distinction between linearly/non-linearly events and synchronous/asynchronous emission of reports the interested reader is encouraged to consult \cite{Afantenos.06:PhD:English,Afantenos&al.05:RANLP,Afantenos&al.07:JIIS,Afantenos&al.05:NLUCS}.} The preprocessing stage involved tokenization and sentence splitting in the first case study and tokenization, sentence splitting and part-of-speech tagging in the second case study. For the task of the \emph{entities recognition and classification} in the first case the use of simple gazetteer lists proved to be sufficient. In the second case study this was not the case and thus we opted for using what we called a \emph{cascade of classifiers} which contained three levels. At the first level we used a binary classifier which determines whether a textual element in the input text is an instance of an ontology concept or not. At the second level, the classifier takes the instances of the ontology concepts of the previous level and classifies them under the top-level ontology concepts (e.g. \texttt{Person}). Finally at the third level we had a specific classifier for each top-level ontology concept, which classifies the instances in their appropriate sub-concepts; for example, in the \texttt{Person} ontology concept the specialized classifier classifies the instances into \texttt{Offender, Hostage}, etc. For the third stage of the messages' extraction we use in both case studies lexical and semantic features. As lexical features in the first case we used the words of the sentences (excluding low frequency words and stop-words) while in the second case study we used only the verbs and nouns of the sentences as lexical features. As semantic features in the first case study we used the number of the top-level ontology concepts that appear in the sentence, while in the second case study we enriched that with the appearance of certain trigger words in the sentence. Finally, the extraction of the \sdrs is the most straightforward task, since the only thing that is needed is the translation of the relations' specifications into an appropriate algorithm which, once applied to the extracted messages, will provide the relations that connect the messages, effectively thus creating the grid. In Table~\ref{table:stats} we present the statistics of the final messages and \sdrs extraction stages for both case studies.\footnote{For more details, critique of those results and comparison with related work the interested reader is encouraged to consult \cite{Afantenos.06:PhD:English,Afantenos&al.07:JIIS}.}

\begin{table}[htb]
\centering
\begin{tabular}{|c||l|l|}
  \hline
    & Case Study I & Case Study II \\
  \hline\hline
           & Pr : 91.12\% & Pr : 42.96\%\\
  Messages & Rc : 67.79\% & Rc : 35.91\%\\
           & FM : 77.74\% & FM : 39.12\%\\
  \hline
           & Pr : 89.06\% & Pr : 30.66\%\\
  SDRs     & Rc : 39.18\% & Rc : 49.12\%\\
           & FM : 54.42\% & FM : 37.76\%\\
  \hline
\end{tabular}
\caption{Precision, Recall and F-Measure for the extraction of the Messages and \protect\sdrs for both case studies.}\label{table:stats}
\end{table}

The creation of the grid can be considered as completing---as we have previously argued \cite{Afantenos&al.07:JIIS}---the \emph{Document Planning} phase of a typical architecture of an \NLG system \cite{Reiter&Dale2000}. Nevertheless, this graph can prove to be very large and thus the resulting summary can easily exceed the desired compression rate. In the following two sections we will present a brief sketch of a probabilistic model which can operate on the Content Determination stage of the Document Planning phase in order to select the appropriate content so that the compression rate of the summary will be respected.

\section{The White, Grey, and Black Areas of \protect\mds} \label{sec:greyareas}
Not too distant in time from the dawn of Artificial Intelligence in the early 1950's, the first seeds of automatic text summarization appeared with the seminal works of Luhn \cite{Luhn58} and Edmundson \cite{Edmundson69}. Those early works, as well as the works on summarization that would follow in the next decades, were mostly concerned with the creation of summaries from single documents. Most of them were focusing on the verbatim extraction of important textual elements, usually sentences or paragraphs, from the input document in order to create the final summary. The methods used for the identification of the most salient sentences or paragraphs vary from a mixture of locational criteria with statistics \cite{Edmundson69,Luhn58,Paice81} to statistical based graph creation methods \cite{Salton&al97} to \RST based methods \cite{Marcu2000}.

\begin{figure}[htb]
\begin{center}
  \ifpdf
    \includegraphics[scale=0.18]{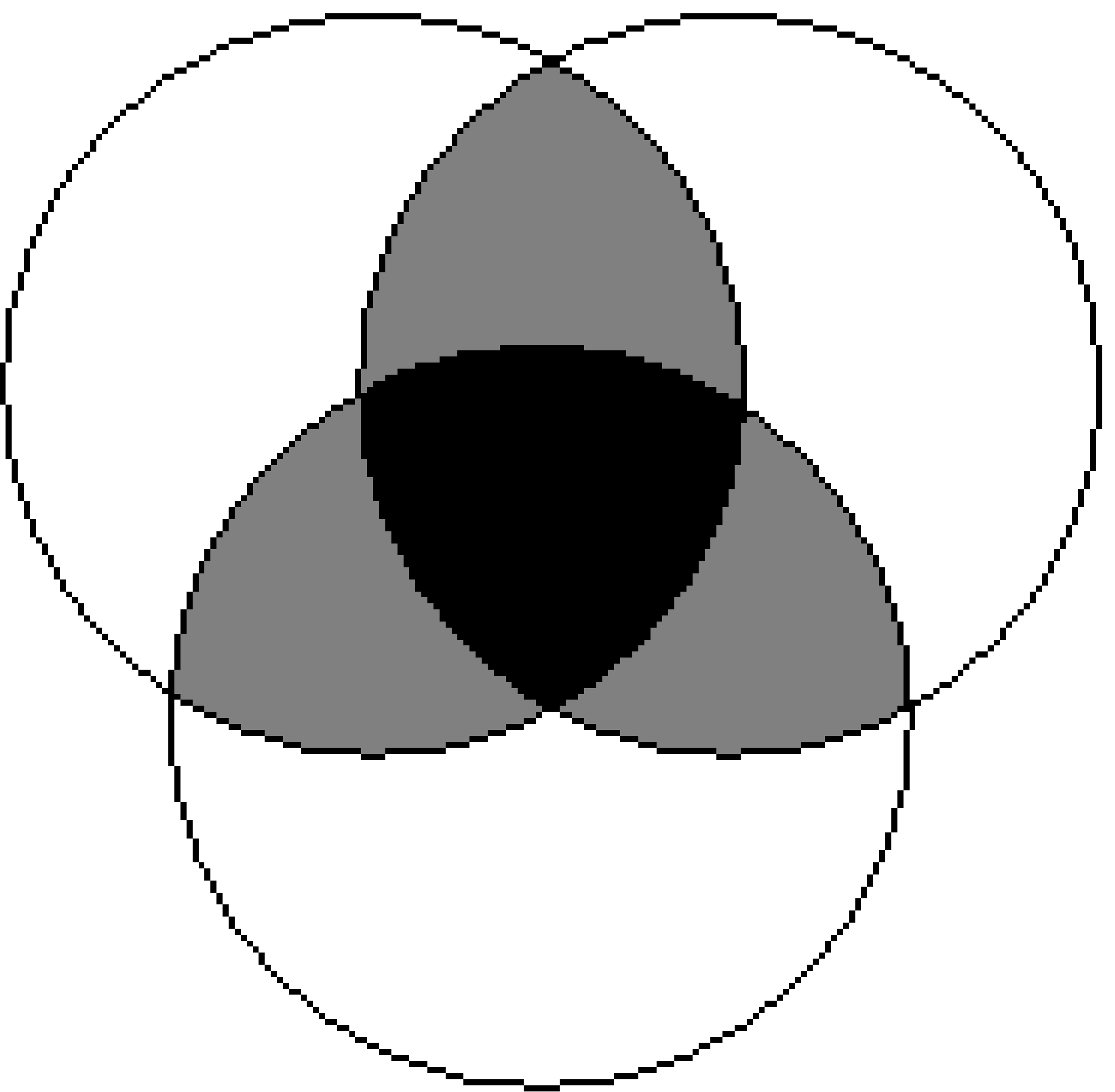}
  \else
    \includegraphics[scale=0.15]{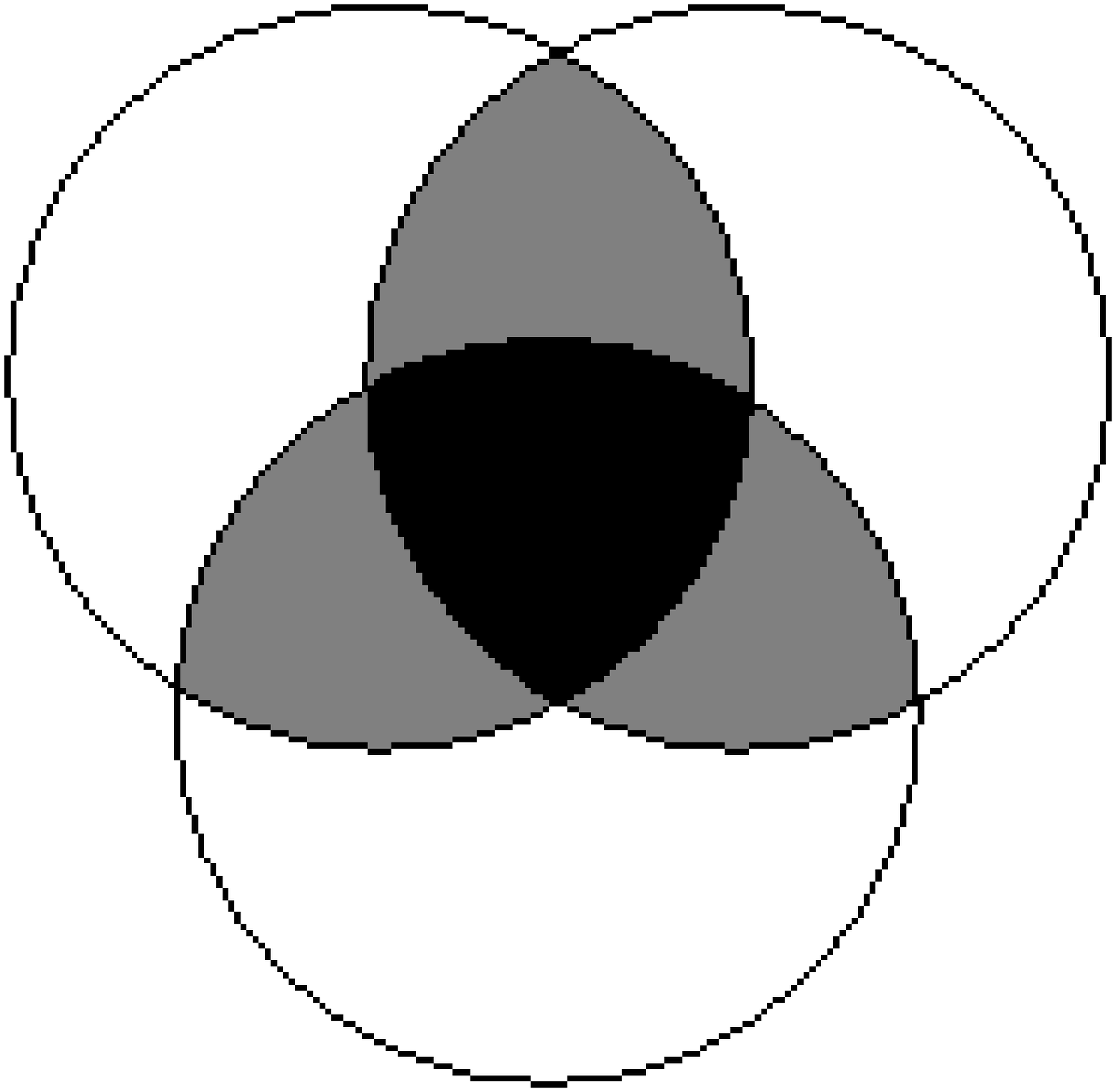}
  \fi
  \caption{Information redundancy and information isolation.}
  \label{fig:information}
\end{center}
\end{figure}

\mdslong would not be actively pursued by researchers up until the mid 1990's, since when it is a quite active area of research.\footnote{For a general overview of summarization the interested reader is encouraged to consult \cite{Mani01}. Mani~\textit{\&}~Maybury \cite{Mani&Maybury99} provide a wonderful collection of papers on summarization spanning most of the research sub-fields of this area. Afantenos~\textit{et~al.}~\cite{Afantenos&al.05:MedicalSurvey} provide an overview as well, focusing mostly on the summarization from medical documents. Finally, \cite{Endres-Niggemeyer98} contains an excellent account of the \emph{cognitive processes} that are involved during the task of single document summarization by professionals, as well a brief overview of the field of summarization.} The main difference that seems to exist between the summarization of a single document and the summarization of multiple (related) documents, seems to be the fact that the ensemble of the related documents, in most of the cases, creates \emph{informational redundancy}, as well as what---for a lack of better term---we will call \emph{informational isolation}. In the case of informational redundancy more than one document contain the same information, while in the case of informational isolation only one document contains a specific piece of information. This is graphically depicted in Figure~\ref{fig:information}, in which each circle represents the information that is contained in a different document. The black and grey areas of the figure represent the information redundancy that exists between the documents. More specifically, the black area represents information which is common to all of the documents, while the grey areas represent information which are common between some articles but not all of them. The white areas, on the other hand, represent what we have called the informational isolation of certain portions of texts, in the sense that the information contained therein is not found anywhere else in the collection of documents.

Of course, one could imagine many more ways in which the circles could be arranged. For example, a circle could be contained inside two other circles, which would imply that the corresponding document is informationally subsumed by the other two. More extreme cases can involve circles arranged in a way that only gray areas exist, which would imply that the documents of the collection are only very loosely related, or cases in which one or more circles are completely white, meaning that the documents which are represented by those circles are completely unrelated with the rest of the documents. Such cases though, one could argue, violate the premises of \mds which require a set of \emph{related documents} that will be informationally condensed by the end of the process.

Despite those extreme cases, it is fair to assume that the configuration depicted in Figure~\ref{fig:information} represents a fairly common situation in most of the \mds scenarios. Of course we have to bare in mind that in most of the cases we will not have just three documents to be summarized, but most possibly many more. This will have the consequence that the grey areas will not have a single shade of greyness but instead they will range from light grey to dark grey depending on the degree of information overlap that will exist between the various sources.

\section{What Should Be Includ\-ed in a Multi-Document Summary of Evolving E\-vents?}\label{sec:model}
Having made the above distinction between the different levels of information overlap, the question that arises at this point is which pieces of information should finally be included in the text that will summarize the multiple documents. The obvious answer to this question would be that such a summary should include the information that are contained in the input documents in decreasing order of their importance, until the length of the summary reaches the required compression rate of the total length of the input documents. In other words, a summary should contain the black areas of Figure~\ref{fig:information}, then the darker to the lighter grey areas, until the length of the summary reaches the required compression rate.

In mathematical terms this can be expressed as follows. If $P(i)$ is the probability that a piece of information will be included in the final summary, then we can claim that:
\[ P(i) = \frac{\sum_{k=1}^nd_{ki}}{n} \]
where $n$ represents the total number of documents, $d_k$ the $k$-th document, and:
\begin{small}
\[
  d_{ki} = \left\{
              \begin{array}{r@{\quad}l}
                1 & \text{if } d_k \text{ contains information } i\\
                0 & \text{if } d_k \text{ does not contain information } i\\
              \end{array} \right.
\]
\end{small}
Additionally, if $c$ is the desirable compression rate, then the final summary $S$ should confront to the following constraint:
\[ length(S) \leq c \sum_{k=1}^nlength(d_k) \]

\subsection{Objections to the Proposed Model for the General Case of \protect\mds}
Now, the above model is really a simplistic one and a host of objections could be raised concerning its usefulness in the general case of \mds, something that we do acknowledge. One could for example claim that the information that will be contained in the black areas will tend to be trivial information, in the sense that they can be characterized as representing ``common knowledge''. This objection can be balanced by two arguments. The first is that the authors of the original documents will most possibly not contain in their articles such common knowledge, unless it is necessary, in which case it might be a good idea to be included in a summary. The second argument is that if the summarization system uses knowledge representation methods---an ontology for example---then such trivial information will tend not to be included in this knowledge representation. Of course, if the system uses purely statistical methods, then the last argument does not hold.

The second objection concerns the white or light grey areas. In the proposed model such areas will have a small probability of being included in the final summary. Nevertheless, it can be argued that under certain circumstances it can be the case that a piece of information which is mentioned only by one or very few sources might turn out to be very important. For example, a prominent source might have an exclusive piece of information that other sources do not have which might prove to be important for inclusion in the final summary. In such case the proposed model, indeed, will fail to include this piece of information in the final summary.

\subsection{Why the Proposed Model Can Be Considered as a Good Starting Point for the Case of \protect\mds for Evolving Events}
The above discussion outlines some of the objections that might arise when the proposed model is applied under the prism of the general case of \mdslong. Despite those objections, we make the claim in this paper that the proposed model can nevertheless be considered as a good starting point for the case of \mdslong of Evolving Events, at least in the framework we have described in Section~\ref{sec:PhD}.

Concerning the first objection---\emph{i.e.} the claim that the same trivial information might be contained in all the documents and thus such trivial information will have a high probability of being included in the final summary---this claim is rebuffed by the nature of the methodology that we have briefly presented in Section~\ref{sec:PhD} and more fully exposed in \cite{Afantenos.06:PhD:English} and \cite{Afantenos&al.07:JIIS}. The use of an ontology and especially the use of the messages guarantee that the system will try to extract information whose nature, we know beforehand, will be non-trivial. Of course, this beneficial situation has its drawbacks as well. As we have argued in \cite{Afantenos&al.07:JIIS} the creation of the ontology and the specifications of the messages require a considerable amount of human labor. Nevertheless, in Section~9 of \cite{Afantenos&al.07:JIIS} we present specific propositions of how this problem can be alleviated.

Let us now come to the second objection. According to this objection, it can be the case that a piece of information while mentioned by only one or very few sources (which implies that this piece of information stands very few chances of being included in the summary, according to the proposed model of Section~\ref{sec:model}) it might nevertheless be mentioned by a prominent source and thus ought finally to be included in the summary. Although this could be the case, we have to note as well that such prominent sources are usually highly influential ones as well. This has the implication that if a piece of information---which was initially exclusively mentioned by one source only---is indeed an important one for the description of the event's evolution, then, almost surely, the rest of the sources will sooner or later follow the initial source in mentioning this information. Thus what was initially a light grey area, according to the discussion of Section~\ref{sec:greyareas}, will tend to become darker grey, or even black, as time goes by, if indeed the mentioned piece of information is important and thus worthy of inclusion in the final summary of the event's evolution.

This leaves us with the conclusion that the afore presented model can indeed serve as a nice starting point for the Content Determination stage, in the case that the grid contains more messages than the required compression rate requires.\footnote{It would be fair to mention that the above conclusion is  valid in the case that we do have the final set of documents which describe the evolution of the event. In case that the evolution is still on-going and this set is not yet finalized, then it might be the case that the second objection  still holds.}

\section{Conclusions}\label{sec:concl}
In \cite{Afantenos.06:PhD:English} and \cite{Afantenos&al.07:JIIS}
we thoroughly presented a methodology (and applied it in two different case studies) which aims towards the creation of summaries from descriptions of evolving events which are emitted from multiple sources. The end result of this methodology is the computational extraction of a structure, which we called a grid. This structure is a directed acyclic graph (DAG) whose nodes are the messages extracted from the input documents and whose vertices are the \sdrslong that connect those messages. The creation of the grid, as we have argued, completes the Document Planning stage of a typical \NLG architecture.

Nevertheless, it can be the case that the created grid can prove to be large enough in order for the final summary to exceed the required compression rate. In this paper we have presented a probabilistic model which can be applied to the Content Determination stage of the Document Planning phase. The application of that model\footnote{Although the probabilistic model presented in Section~\ref{sec:model} talks about ``pieces of information'' the substitution of this abstract notion with the more concrete concept of \emph{messages} makes that model ready for use in our methodology.} to the extracted grid will have the effect of creating a \emph{subset} of the original grid (a \emph{sub-grid} in other words) which will contain just the messages that confront to this model as well as the \sdrs that connect \emph{only} the selected messages.

From the discussion in this paper, as well as from the general literature in the area of \mdslong, we can conclude that the identification of similarities and differences is an essential component for any \mds system. Digressing a little bit at this point, we would like to note that spotting similarities between even disparate situations or objects, is something that human beings effortlessly and continuously perform all the time, and thus the study of this phenomenon is of paramount importance for the understanding of the human cognitive functioning. The mechanism of identifying ``sameness''---despite its subtlety \cite{French.95}---is an essential component for the task of analogy-making which lies at the core of cognition as \cite{Hofstadter.01} has claimed.

Closing this digression on the fascinating topic of analogy-making\footnote{The interested reader is encouraged to consult \cite{French.95,Gentner&al.01} and \cite{Mitchel.93} for more information on this topic.} we would like to note that with respect to \mds, to the best of our knowledge, there are no empirical studies as to how human beings proceed in order to create a summary from multiple documents---be they documents that describe evolving events, or not. We do not even have sufficient corpora of summaries from multiple documents which will provide us with an insight as to what can be considered a ``good'' multi-document summary. This comes in contrast with the area of \sds in which, of course, we do have such corpora. Moreover, in \sds we do have at least one substantial research from the perspective of Cognitive Science \cite{Endres-Niggemeyer98} which studies the cognitive mechanisms---or ``strategies'' as they are called in that book---of professional summarizers during the process of creating a summary from a single document. It is our personal belief that the performance of more such studies from the cognitive science perspective, for \sds and \mds alike, will be beneficial for the advancement of our understanding not only of how we do create summaries, but for the understanding of how we spot similarities and differences; a task which lies at the heart of analogy-making as well.

\bibliographystyle{abbrv}
\begin{scriptsize}
\bibliography{RANLP07-minimal.bib}
\end{scriptsize}

\end{document}